\documentclass{article}
\usepackage{arxiv}
\usepackage{amsfonts}       
\usepackage{nicefrac}
\usepackage[utf8]{inputenc} 
\usepackage[T1]{fontenc}  
\usepackage{microtype} 
\usepackage{url} 
\usepackage{booktabs}
\usepackage[utf8]{inputenc}
\usepackage{subcaption}
\usepackage{csquotes}
\usepackage{xcolor}
\usepackage{graphicx}
\usepackage{amsmath,amssymb,amsthm}
\usepackage{algorithm}
\usepackage[noend]{algorithmic}
\usepackage{float}
\usepackage{soul}
\usepackage{natbib}
\usepackage{tikz}
\usepackage{pgfplots}
\sethlcolor{lightgray}

\DeclareMathOperator*{\argmax}{arg\,max}
\DeclareMathOperator*{\argmin}{arg\,min}
\pgfplotsset{compat=1.17}

\title{Co-Evolutionary Diversity Optimisation for the Traveling Thief Problem}
\author{ Adel Nikfarjam \\
Optimisation and Logistics\\School of Computer Science\\The University of Adelaide\\
  \texttt{adel.nikfarjam@adelaide.edu.au} \\
  \And
  Aneta Neumann\\
Optimisation and Logistics\\School of Computer Science\\The University of Adelaide\\
  \texttt{aneta.neumann@adelaide.edu.au} \\
  \And
  Jakob Bossek\\
  AI Methodology\\ Dept. of Computer Science\\ RWTH Aachen University \\
  \texttt{bossek@aim.rwth-aachen.de}
  \And
   Frank Neumann \\
Optimisation and Logistics\\School of Computer Science\\The University of Adelaide\\
  \texttt{frank.neumann@adelaide.edu.au} \\}

\begin{document}
\maketitle
\begin{abstract}
    Recently different evolutionary computation approaches have been developed that generate sets of high quality diverse solutions for a given optimisation problem. Many studies have considered diversity 1) as a mean to explore niches in behavioural space (quality diversity) or 2) to increase the structural differences of solutions (evolutionary diversity optimisation). In this study, we introduce a co-evolutionary algorithm to simultaneously explore the two spaces for the multi-component traveling thief problem. The results show the capability of the co-evolutionary algorithm to achieve significantly higher diversity compared to the baseline evolutionary diversity algorithms from the the literature.
\end{abstract}


\section{Introduction}

Diversity has gained increasing attention in the evolutionary computation community in recent years. In classical optimisation problems, researchers seek a single solution that results in an optimal value for an objective function, generally subject to a set of constraints. The importance of having a diverse set of solutions has been highlighted in several studies~\cite{neumann2019evolutionary,NikfarjamB0N21b}. Having such a set of solutions provides researchers with 1) invaluable information about the solution space, 2) robustness against imperfect modelling and minor changes in problems and 3) different alternatives to involve (personal) interests in decision-making. Traditionally, diversity is seen as exploring niches in the fitness space. However, two paradigms, namely quality diversity~(QD) and evolutionary diversity optimisation~(EDO), have been formed in recent years.

QD achieves diversity in exploring niches in behavioural space. QD maximises the quality of a set of solutions that differ in a few predefined features. Such a set of solutions can aid in the grasp of the high-quality solutions' behaviour in the feature space. QD has a root in novelty search, where researchers seek solutions with new behaviour without considering their quality~\cite{LehmanS11}. For the first time, a mechanism is introduced in~\cite{CullyM13} to keep best-performing solutions whereby, searching for unique behaviours. At the same time, the MAP-Elites framework was introduced in~\cite{CluneML13} to plot the distribution of high-performing solutions over a behavioural space. It has been shown that MAP-Elites is efficient in evolving behavioural repertoires. Later, the problem of computing a set of best-performing solutions differing in terms of some behavioural features is formulated and named QD in~\cite{PughSSS15,PughSS16}. 

In contrast to QD, the goal of EDO is to explicitly maximise the structural diversity of a set of solutions that all have a desirable minimum quality. This approach was first introduced in~\cite{ulrich2011maximizing} in the context of continuous optimisation. Later, EDO was adopted to generate images and benchmark instances for the traveling salesperson problem~(TSP)~\cite{abs-2011-05081,doi:10.1162/evcoa00274}. Star-discrepancy and performance indicators from multi-objective evolutionary optimisation were adopted to achieve the same goals in~\cite{neumann2018discrepancy,neumann2019evolutionary}. In recent years EDO was studied in the context of well-known combinatorial optimisation problems, such as the quadratic assignment problem~\cite{DoGN021}, the minimum spanning tree problem~\cite{Bossek021tree}, the knapsack problem~\cite{BossekN021KP}, and the optimisation of monotone sub-modular functions~\cite{NeumannB021}.
Distance-based diversity measures and entropy have been incorporated into EDO to evolve diverse sets of high-quality solutions for the TSP~\cite{viet2020evolving,NikfarjamBN021a}. Nikfarjam et al.~\cite{NikfarjamB0N21b} introduced an EAX-based crossover focusing on structural diversification of TSP solutions. Most recently, Neumann et al.~\cite{AnetaCo-EA} introduced a co-evolutionary algorithm to find Pareto-front for bi-objective optimisation problem and simultaneously evolve another population to maximise structural diversity.  

In this paper, we introduce a co-evolutionary algorithm~(Co-EA) to compute two sets of solutions simultaneously; one employs the QD concept and the other evolves towards EDO. We consider the traveling thief problem~(TTP) as a well-studied multi-component optimisation problem. QD and EDO have separately been studied in the context of TTP in~\cite{NikfarjamMap} and \cite{NikTTPEDO}, respectively. However, the Co-EA has several advantages:
\begin{itemize}
    \item QD provides researchers with invaluable information about the distribution of best-performing solutions in behavioural space and enables decision-makers to select the best solution having their desirable behaviour. On the other hand, EDO provides us with robustness against imperfect modelling and minor changes in problems. we can benefit from both paradigms by using the Co-EA.
    \item Optimal or close-to optimal solutions are required in most EDO studies for initialization. The Co-EA eliminates this restriction.
    \item We expect the Co-EA brings about better results, especially in terms of structural diversity since the previous frameworks are built upon a single solution (the optimal solution). The Co-EA eliminates this drawback.
    \item The Co-EA benefits from a self-adaptation method to tune and adjust some hyper-parameters during the search improving the results meaningfully.
\end{itemize}
 
The remainder of the paper is structured as follows. We formally define the TTP and diversity in Section \ref{Sec:prob_def}. The Co-EA is introduced in  Section~\ref{Sec:alg}. We conduct comprehensive experimental investigation to evaluate Co-EA in Section \ref{Sec:EXP}. Finally, we finish with concluding remarks. 
 
\section{Preliminaries}
\label{Sec:prob_def}

In this section, we introduce the traveling thief problem and outline different diversity optimisation approaches established for this problem.

\subsection{The Traveling Thief Problem}

The traveling thief problem~(TTP) is a multi-component combinatorial optimization problem. I.~e., it is a combination of the classic traveling salesperson problem~(TSP) and the knapsack problem~(KP). The TSP is defined on a graph $G=(V, E)$ with a node set $V$ of size $n$ and a set of pairwise edges $E$ between the nodes, respectively. Each edge, $e = (u, v) \in E$ is associated with a non-negative distance $d(e)$. In the TSP, the objective is to compute a tour/permutation $x : V \to V$ which minimizes the objective function
$$f (x) = d(x(n),x(1)) + \sum_{i=1}^{n-1} d(x(i),x(i+1)).$$
The KP is defined on a set of items $I$ with $m := |I|$. Each item $i \in I$ has a profit $p_i$ and a weight $w_i$. 
The goal is to determine a selection of items, in the following encoded as a binary vector $y = (y_1, \ldots, y_m) \in \{0,1\}$, that maximises the profit, while the selected items' total weight does not exceed the capacity $W>0$ of the knapsack: 
\begin{align*}
& g (y) = \sum_{j=1}^{m} p_j y_j
\text{ s.~t. } \sum_{j=1}^{m} w_j y_j \leq W.
\end{align*}
Here, $y_j= 1$ if the $j$th item is included in the selection and $y_j=0$ otherwise.

The TTP is defined on a graph $G$ and a set of items $I$. Each node $i$ except the first one includes a set of items $M_i \subseteq I$. In TTP, a thief visits each city exactly once and picks some items into the knapsack. A rent $R$ is to be paid for the knapsack per time unit, and the speed of thief non-linearly depends on the weight $W_{x_i}$ of selected items so far. Here, the objective is to find a solution $p=(x, y)$ including a tour $x$ and a packing list (the selection of items) $y$ that maximises the following function subject to the knapsack capacity:
\begin{align*}
& z(p) = g(y) - R \left( \frac{d(x(n), x(1))}{\nu_{\max}-\nu W_{x_n}} + \sum_{i=1}^{n-1} \frac{d(x(i) ,x(i+1))}{\nu_{\max}-\nu W_{x_i}} \right)
\text{ s.~t. } \sum_{j=1}^{m} w_j y_j \leq W.
\end{align*}
where $\nu_{\max}$ and $\nu_{\min}$ are the maximal and minimal traveling speed, and $\nu = \frac{\nu_{\max}-\nu_{\min}}{W}$.

\subsection{Diversity Optimisation}
This study simultaneously investigates QD and EDO in the context of the TTP. For this purpose, two populations $P_1$ and $P_2$ co-evolve. $P_1$ explores niches in the behavioural space and the $P_2$ maximises its structural diversity subject to a quality constraint.
In QD, a behavioural descriptor~(BD) is defined to determine to which part of the behavioural space a solution belongs. In line with \cite{NikfarjamMap}, we consider the length of tours $f(x)$, and the profit of selected items $g(y)$, to serve as the BD. To explore niches in the behavioural space, we propose a MAP-Elites-based approach in the next section.

For maximising structural diversity, we first require a measure to determine the diversity. For this purpose, we employ the entropy-based diversity measure in \cite{NikTTPEDO}. Let $E(P_2)$ and $I(P_2)$ denote the set of edges and items included in population $P_2$. The structural entropy of $P_2$ defines on two segments, the frequency of edges and items included in $E(P_2)$ and $I(P_2)$, respectively. let name these two segments edge and item entropy and denote them by $H_e$ and $H_i$. $H_e$ and $H_i$ are calculated as
\begin{align*}
H_{e}(P_2) = \sum_{e \in E(P_2)} h(e) \text{ with } h(e)=-\left(\frac{f(e)}{\sum_{e \in I} f(e)}\right) \cdot \ln{\left(\frac{f(e)}{\sum_{e \in I} f(e)}\right)}
\end{align*}
and
\begin{align*}
H_i(P_2) = \sum_{i \in I(P_2)} h(i) \text{ with } h(i)=-\left(\frac{f(i)}{\sum_{i \in I} f(i)}\right) \cdot \ln{\left(\frac{f(i)}{\sum_{i \in I} f(i)}\right)}
\end{align*}
where $h(e)$ and $h(i)$ denote the contribution of edge $e$ and item $i$ to the entropy of $P_2$, respectively. Also, the terms $f(e)$ and $f(i)$ encode the number of solutions in $P_2$ that include $e$ and $i$. It has been shown that $\sum_{e \in E(P_2)} f(e) = 2n\mu$ in~\cite{NikfarjamBN021a} , where $\mu = |P_2|$, while the number of selected items in $P_2$ can fluctuate. The overall entropy of $P_2$ is calculated by summation
$$H(P_2) = H_e(P_2) + H_i(P_2).$$
$P_2$ evolves towards maximisation of $H(P_2)$ subject to $z(p) \geq z_{\min}$ for all $p \in P_2$. Overall, we maximise the solutions' quality and their diversity in the feature-space through $P_1$, while we utilise $P_2$ to maximises the structural diversity.    

\section{Co-Evolutionary Algorithm}
\label{Sec:alg}

This section presents a co-evolutionary algorithm -- outlined in Algorithm~\ref{alg:map} -- to simultaneously tackle QD and EDO problems in the context of TTP. The algorithm involves two populations $P_1$ and $P_2$, employing MAP-Elite-based and EDO-based selection procedures.

\subsection{Parent Selection and Operators}
A bi-level optimisation procedure is employed to generate offspring. A new tour is generated by crossover at the first level; then, $(1+1)$~EA is run to optimise the packing list for the tour. The crossover is the only bridge between $P_1$ and $P_2$. For the first parent we first select $P_1$ or $P_2$ uniformly at random. Then, one individual, $p_1(x_1, y_1)$ is selected again uniformly at random from the chosen population; the same procedure is repeated for the selection of the second parent $p_2(x_2, y_2)$. To generate a new solution $p'(x', y')$ from $p_1(x_1, y_1)$ and $p_2(x_2, y_2)$, a new tour $x' \gets crossover(x_1, x_2)$ is first generated by EAX-1AB crossover. Edge-assembly crossover~(EAX) is a high-performing operator and yields strong results in solving TSP. Nikfarjam et al.~\cite{NikfarjamMap} showed that the crossover performs decently for the TTP as well. 

EAX-1AB includes three steps: It starts with generating a so-called AB-Cycle of edges by alternatively selecting the edges from parent one and parent two. Next, an intermediate solution is formed. Having the first parent's edges copied to the offspring, we delete parent one's edges included in the AB-cycle and add the rest of edges in the AB-cycle. In this stage, we can have either a complete tour or a number of sub-tours. In latter case, we connect all the sub-tours one by one stating from the sub-tour with minimum number of edges. For connecting two sub-tours, we discard one edge from each sub-tour and add two new edges, a $4$-tuple of edges. The $4$-tuple is selected by following local search by choosing
$$(e_1, e_2, e_3, e_4) = \argmin \{-d(e_1)-d(e_2)+d(e_3)+d(e_4)\}.$$
Note that if $E(t)$ and $E(r)$ respectively show the set of edges of the the intermediate solution $t$ and the sub-tour $r$, $e_1\in E(r)$ $e_2 \in E(t) \setminus E(r)$. We refer interested readers to~\cite{nagata2013powerful} for details on the implementation of the crossover.

Then, an internal $(1+1)$~EA is started to optimise a packing list $y'$ for the new tour $x'$ and form a complete TTP solution $p'(x', y')$. The new solution first inherits the first parent's packing list, $y' \gets y_1$. Next, a new packing list is generated by standard bit-flip mutation ($y'' \gets mutation(y')$). If $z(x',y'') > z(x', y')$, the new packing list is replaced with old one, $y' \gets y''$. These steps repeats until an internal termination criterion for the $(1+1)$~EA is met. The process of generating a new solution $p'(x',y')$ is complete here, and we can ascend to survival selection.     

\subsection{Survival Selection Procedures}

In MAP-elites, solutions with similar BD compete, and usually, the best solution survives to the next generation. To formally define the similarity and tolerance of acceptable differences in BD, the behavioural space is split into into a discrete grid, where each solution belongs to only one cell. Only the solution with the highest objective value is kept in a cell in survival selection. The map not only contributes to the grasp of the high-quality solutions' behaviour but also does maintain the diversity of the population and aids to avoid premature convergence. 

In this study, we discretize the behavioural space in the same way~\cite{NikfarjamMap} did. They claimed that it is beneficial for the computational costs if we focus on a promising portion of behavioural space. In TTP, solely solving either TSP or KP is insufficient to compute a high-quality TTP solution. However, a solution $p(x, y)$ should score fairly good in both $f(x)$ and $g(y)$ in order to result in a high TTP value $z(p)$. Thus, we limit the behavioural space to the neighbourhood close to optimal/near-optimal values of the TSP and the KP sub-problems. In other words, a solution $p(x, y)$ should result in $f(x) \in [f^*, (1+\alpha_1)\cdot f^*]$ and $g(y)\in[(1-\alpha_2)\cdot g^*, g^*]$. Note that $f^*$ and $g^*$ are  optimal/near-optimal values of the TSP and the KP sub-problems, and $\alpha_1$ and $\alpha_2$ are acceptable thresholds to $f^*$ and $g^*$, respectively. We obtain $f^*$ and $g^*$ by EAX~\cite{nagata2013powerful} and dynamic programming~\cite{DBLP:Toth80}. Next, We discretize the space into a grid of size $\delta_1 \times \delta_2$. Cell $(i,j)$, $1 \leq i \leq \delta_1$, $1 \leq j \leq \delta_2$ contains the best solution, with 
$$f(x) \in \left[f^*+(i-1)\cdot\left(\frac{\alpha_1 f^*}{\delta_1}\right), f^*+i \cdot \left(\frac{\alpha_1 f^*}{\delta_1}\right)\right]$$
and 
$$g(y)\in\left[(1-\alpha_2)\cdot g^*+(j-1)\cdot\left(\frac{\alpha_2 g^*}{\delta_2}\right), (1-\alpha_2)\cdot g^*+j\cdot\left(\frac{\alpha_2 g^*}{\delta_2}\right)\right].$$

After generating a new solution $p'(x', y')$, we find the cell corresponding with its BD ($f(x'), g(y')$); if the cell is empty, $p'$ is added to the cell. Otherwise, the solution with highest TTP value is kept in the cell.

Having defined the survival selection of $P_1$, we now look at $P_2$'s survival selection based on EDO. We add $p'(x', y')$ to $P_2$ if the quality criterion is met, i.~e., $z(p')\geq z_{\min}$. If $|P_2| = \mu+1$, a solution with the least contribution to $H(P)$ will be discarded.

\subsection{Initialisation}

Population $P_1$ only accepts solutions with fairly high BDs ($f(x), g(y)$), and there is a quality constraint for $P_2$. Random solutions are unlikely to have these characteristics. As mentioned, we use the GA in~\cite{nagata2013powerful} to obtain $f^*$; since the GA is a population-based algorithm, we can derive the tours in the final population resulting in a fairly good TSP score. Afterwards, we run the $(1+1)$~EA described above to compute a high-quality packing list for the tours. These packing lists also bring about a high KP score that allows us to populate $P_1$. Depending on the quality constraint $z_{\min}$, the initial solutions may not meet the quality constraint. Thus, it is likely that we have to initialize the algorithm with only $P_1$ until the solutions comply with the quality constraint; then, we can start to populate $P_2$. Note that both parents are selected from $P_1$ while $P_2$ is still empty. We stress that in most previous EDO-studies an optimal (or near-optimal solution) was required to be known a-priori and for initialization. In the proposed Co-EA, this strong requirement is no longer necessary. 

\begin{algorithm}[t!]
\begin{algorithmic}[1]
\STATE Find the optimal/near-optimal values of the TSP and the KP by algorithms in~\cite{nagata2013powerful,DBLP:Toth80}, respectively. 
\STATE Generate an empty map and populate it with the initialising procedure. 

\WHILE{termination criterion is not met}

\STATE Select two individuals based on the parent selection procedure and generate offspring by EAX and $(1+1)$~EA.
\IF{The offspring's TSP and the KP scores are within $\alpha_1$, and $\alpha_2$ thresholds to the optimal values of BD.}
\STATE Find the corresponding cell to the TSP and the KP scores in the QD map.
\IF{The cell is empty}
\STATE Store the offspring in the cell.
\ELSE
\STATE Compare the offspring and the individual occupying the cell and store the best individual in terms of TTP score in the cell.
\ENDIF
\ENDIF
\IF{The offspring complies with the quality criterion}
\STATE Add the offspring to the EDO population.
\IF{The size of EDO population is equal to $\mu+1$}
\STATE Remove one individual from the EDO population with the least contribution to diversity. \ENDIF
\ENDIF
\ENDWHILE
\end{algorithmic}
\caption{The Co-Evolutionary Diversity Algorithm}
\label{alg:map}
\end{algorithm}
\subsection{Self Adaptation}
Generating offspring includes the internal $(1+1)$~EA to compute a high-quality packing list for the generated tour. In~\cite{NikfarjamMap}, the $(1+1)$~EA is terminated after a fixed number of $t = 2m$ fitness evaluations. However, improving the quality of solutions is easier in the beginning and gets more difficult as the search goes on. Thus, we adopt a similar self-adaptation method proposed in~\cite{doerr2015optimal,neumann2017adapt} to adjust $t$ during the search. Let $Z= \argmax_{p \in P_1}\{z(p)\}$. Success defines an increase in~$Z$. We discretize the search to intervals of $u$ fitness evaluations. An interval is successful if $Z$ increases; otherwise it is a failure. We reset $t$ after each interval; $t$ decreases if $Z$ increases during the last interval. Otherwise, $t$ increases to give the internal $(1+1)$~EA more budget in the hope of finding better packing lists and better TTP solutions.
Here, we set $t = \gamma m$ where $\gamma$ can take any value in $[\gamma_{\min}, \gamma_{\max}]$. We set
\begin{align*}
    \gamma := \max \{\gamma \cdot F_1, \gamma_{\min}\}
    \text{ and }
    \gamma := \min \{\gamma \cdot F_2, \gamma_{\max}\}
\end{align*}
in case of success and failure respectively. In our experiments, we use $F_1 = 0.5$, $F_2 = 1.2$, $\gamma_{\min} = 1$, $\gamma_{\max} = 10$, and $u = 2000m$ based on preliminary experiments. We refer to this method as $Gamma_1$.

Moreover, we propose an alternative terminating criterion for the internal $(1+1)$~EA, and denote it $Gamma_2$. Instead of running the $(1+1)$~EA for $t = \gamma m$, we terminate $(1+1)$~EA when it fails in improving the packing list in $t' = \gamma' m$ consecutive fitness evaluations.
$\gamma'$ is updated in the same way as $\gamma$. Based on the preliminary experiments, we set $\gamma'_{\min}$ and $\gamma'_{\max}$ to $0.1$ and $1$, respectively. 
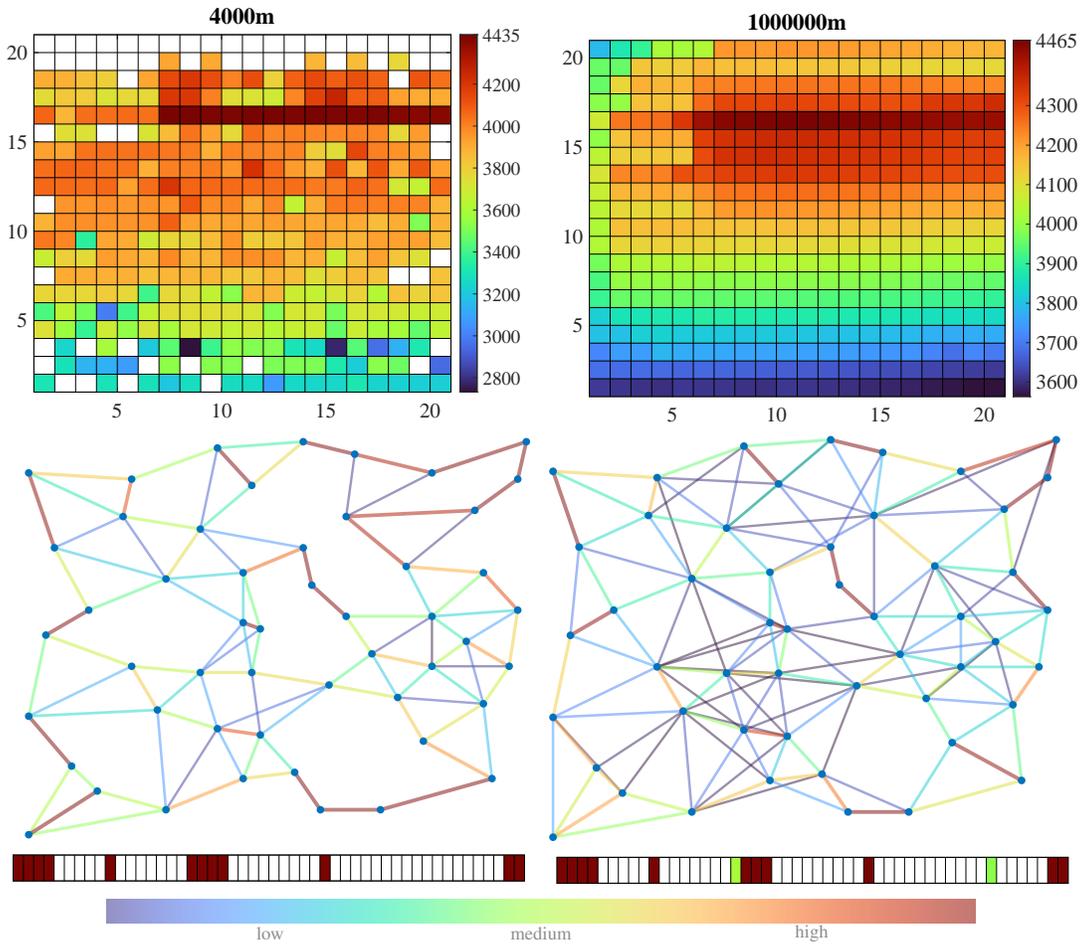
\begin{figure*}
\centering
\includegraphics[width=.42\columnwidth]{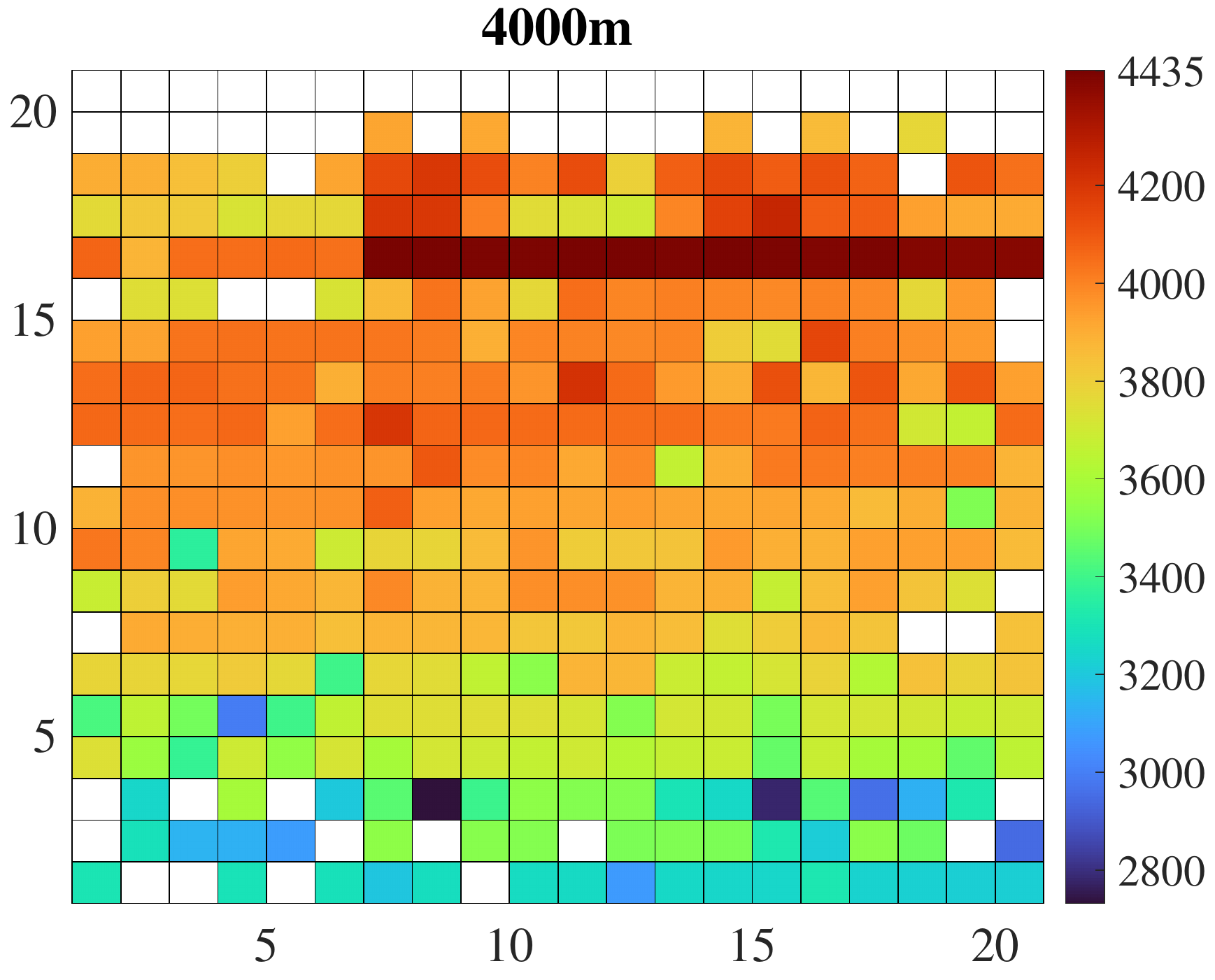}
\hskip10pt
\includegraphics[width=.42\columnwidth]{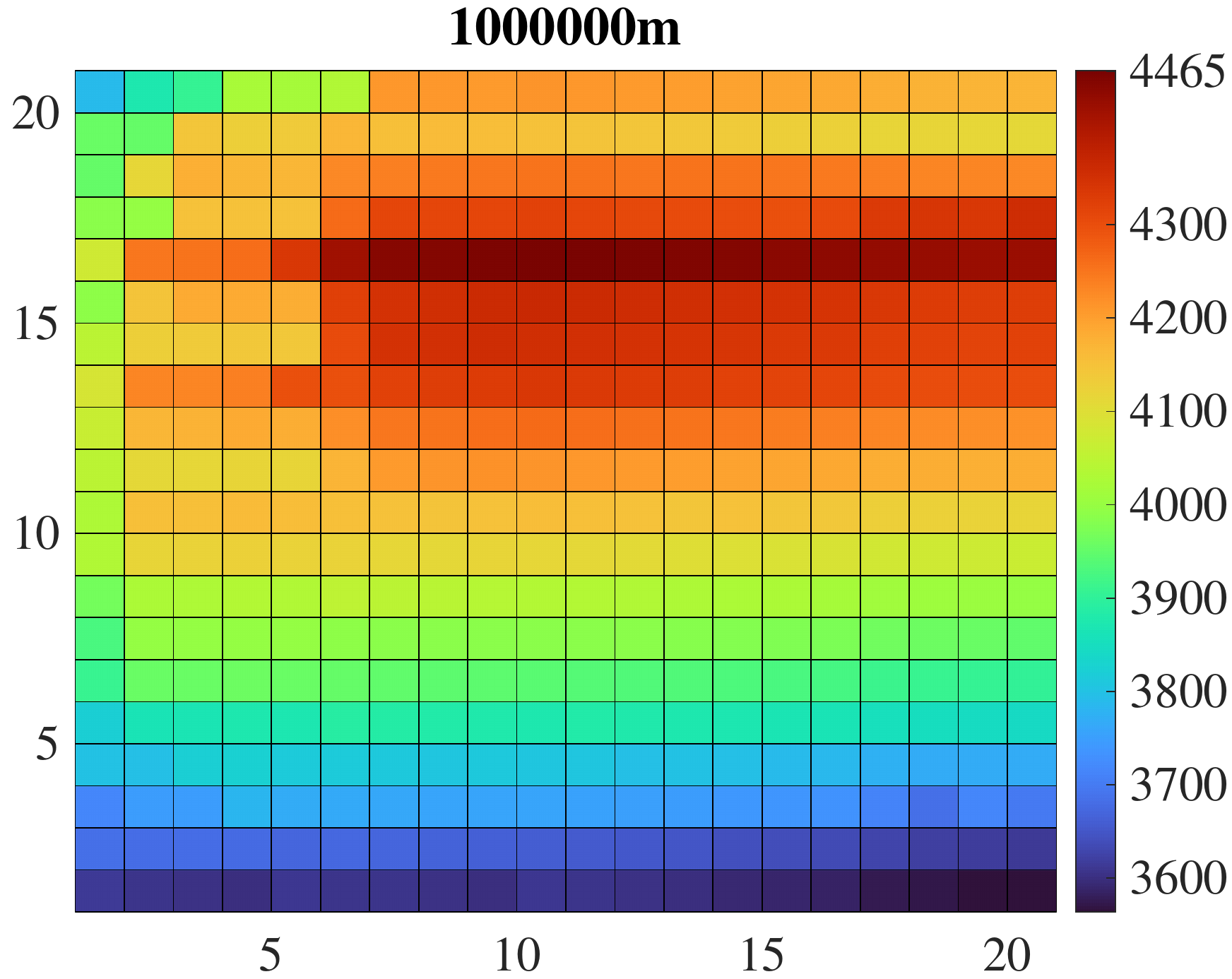}
\hskip10pt
\includegraphics[width=.42\columnwidth]{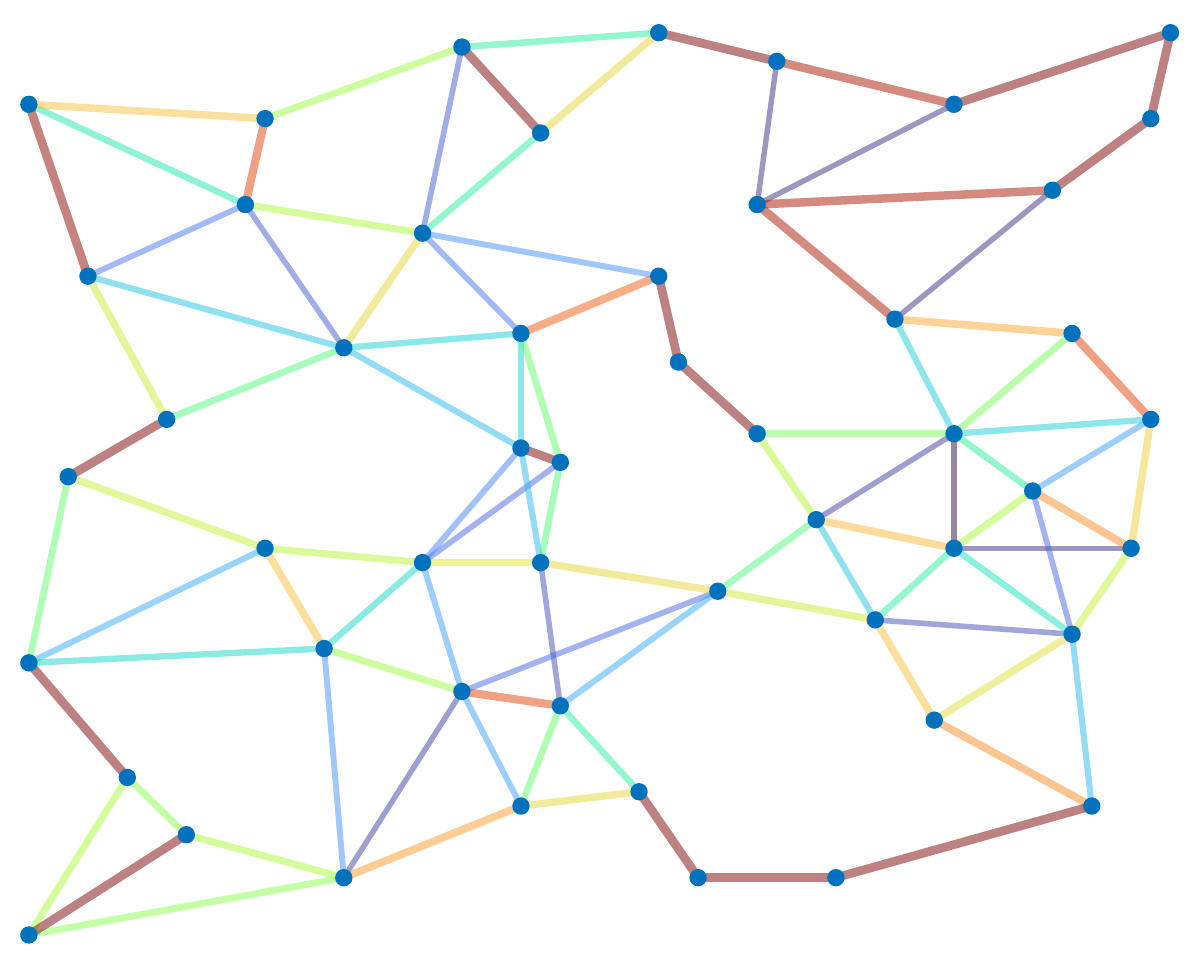}
\includegraphics[width=.42\columnwidth]{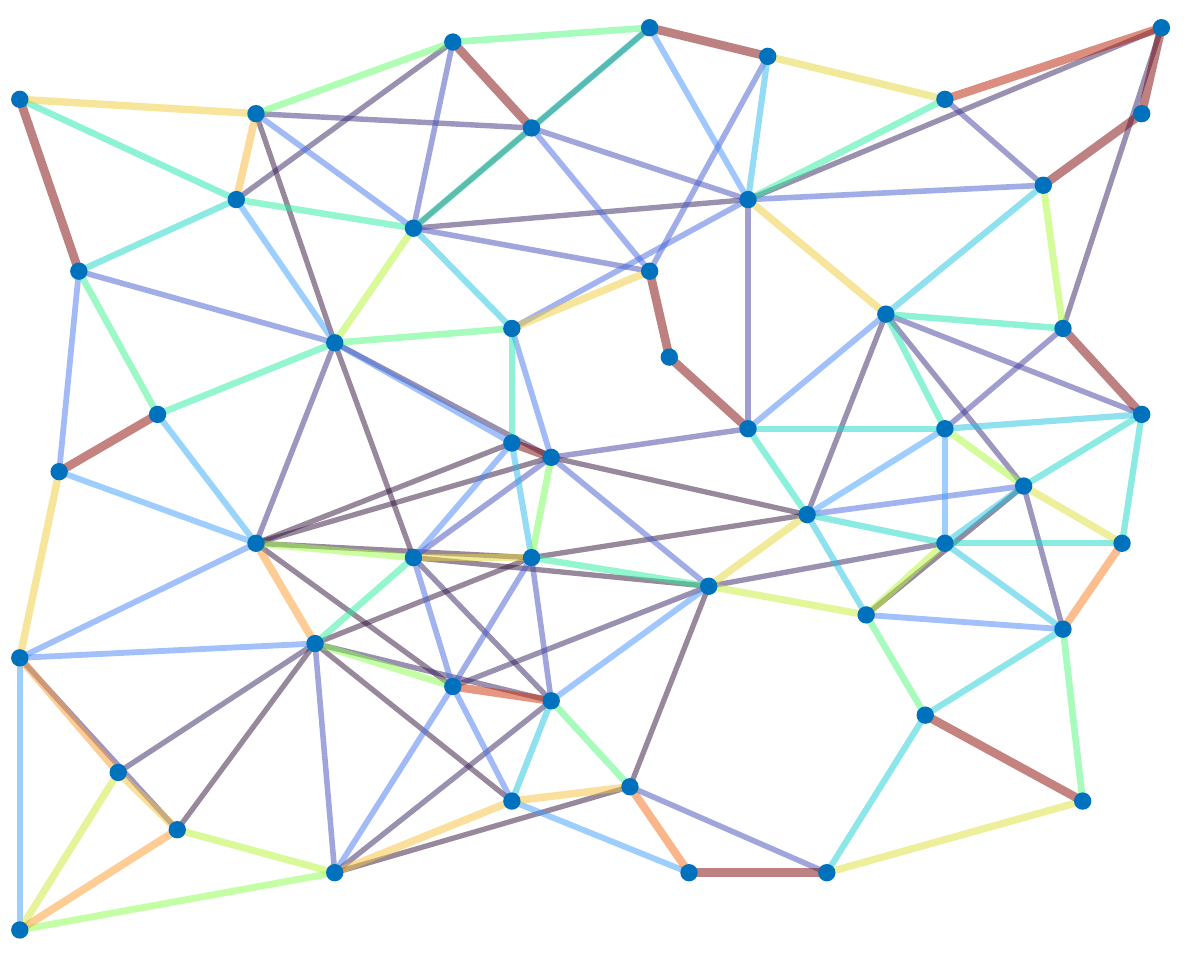}
\hskip10pt
\includegraphics[width=.42\columnwidth]{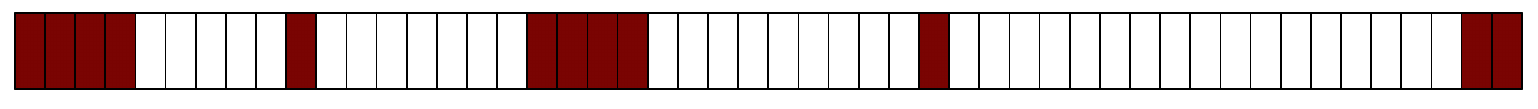}
\hskip6pt
\includegraphics[width=.42\columnwidth]{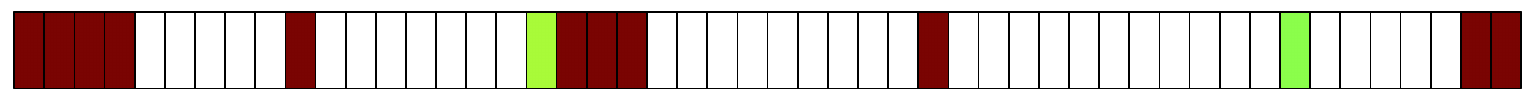}
\begin{tikzpicture}
\node[opacity=.55] (legend) at (0,0) {\includegraphics[width=0.7\columnwidth,trim=20pt 40pt 10pt 0,clip]{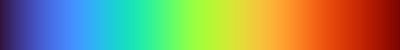}};
\node (low) at (-3.6,-0.3) {\scriptsize{\textcolor{gray!90}{low}}};
\node (medium) at (0,-0.3) {\scriptsize{\textcolor{gray!90}{medium}}};
\node (high) at (3.6,-0.3) {\scriptsize{\textcolor{gray!90}{high}}};
\end{tikzpicture}
\caption{Evolution of $P_1$ and $P_2$ over $4000m$ and $1000000m$ fitness evaluations on instance~$1$ with $\alpha = 2\%$. The first row depicts the distribution of high-quality solutions in the behavioural space ($P_1$). The second and the third rows show the overlay of all edges and items used in exemplary $P_2$, respectively. Edges and items are coloured by their frequency.}
\label{fig:frq+map}
\end{figure*}
\begin{figure*}
\centering
\includegraphics[width=.42\columnwidth]{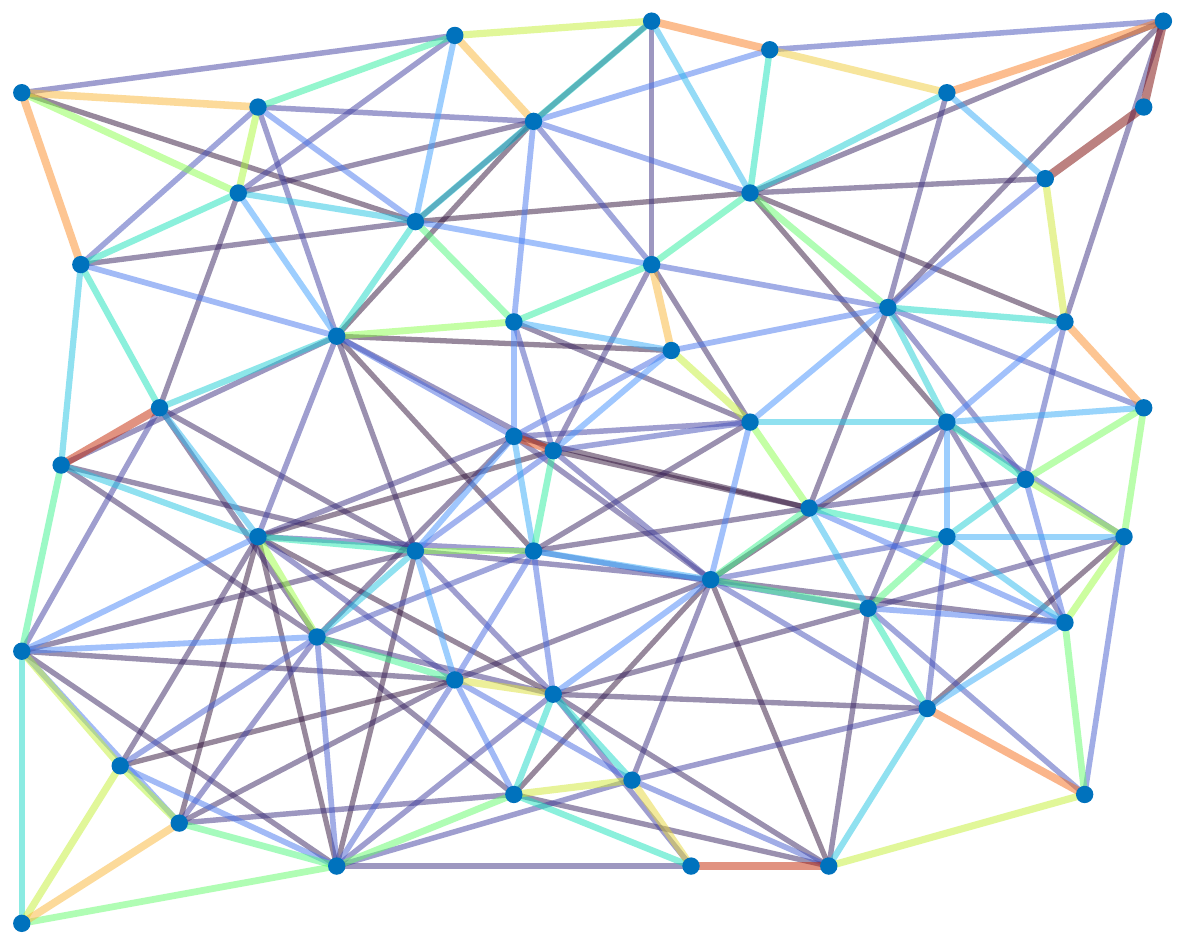}
\includegraphics[width=.42\columnwidth]{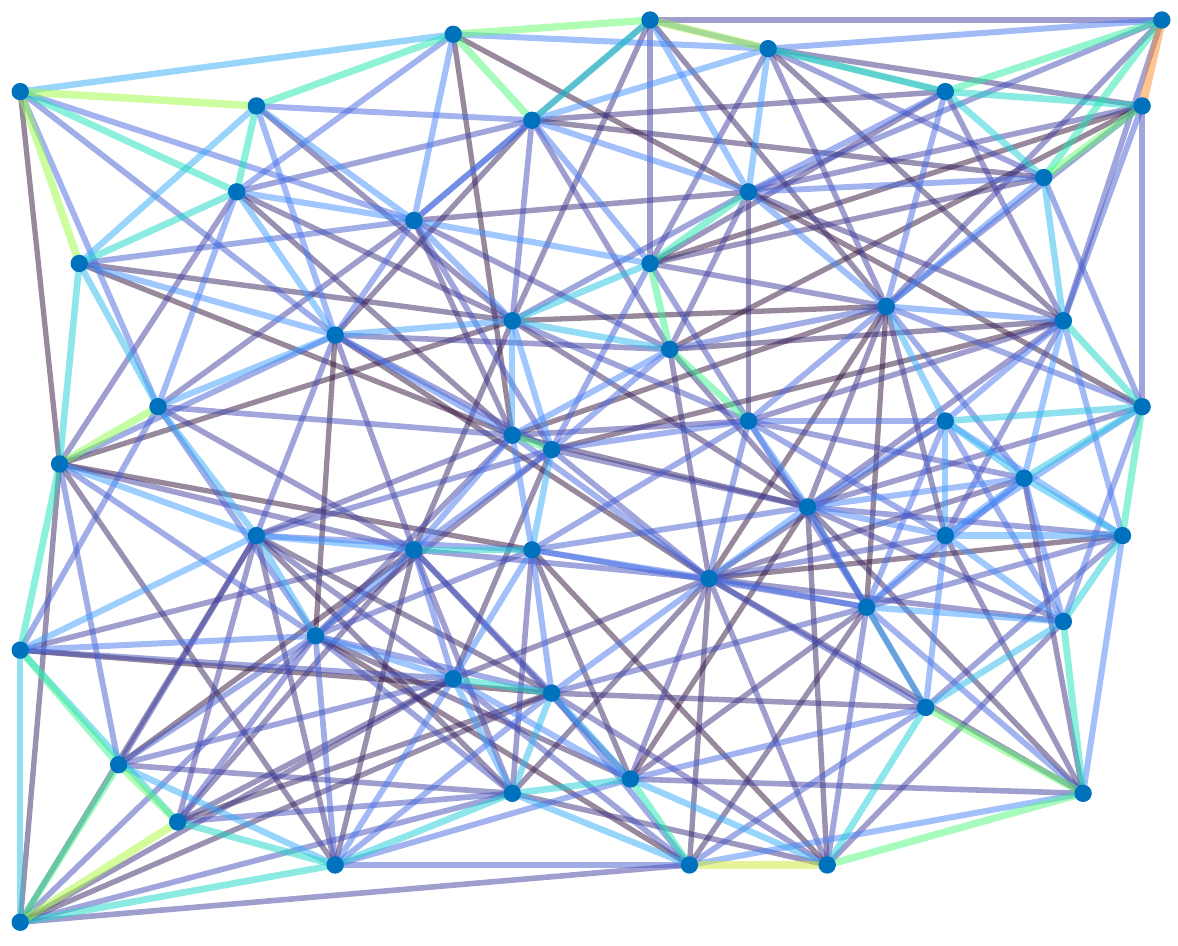}
\hskip10pt
\includegraphics[width=.42\columnwidth]{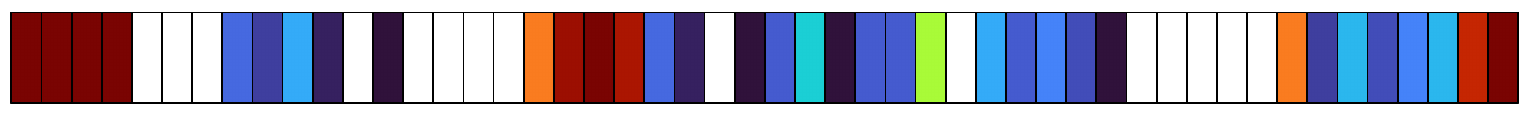}
\includegraphics[width=.42\columnwidth]{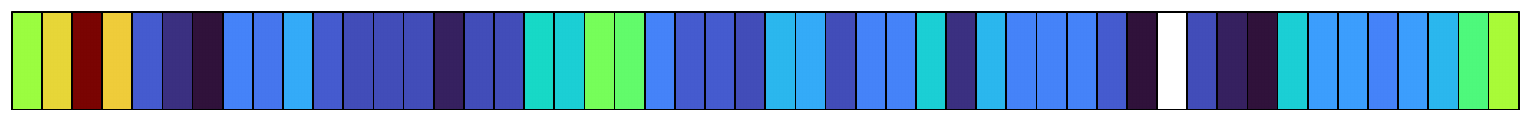}
\begin{tikzpicture}
\node[opacity=.55] (legend) at (0,0) {\includegraphics[width=0.7\columnwidth,trim=20pt 40pt 10pt 0,clip]{turbo_palette_legend.png}};
\node (low) at (-3.6,-0.3) {\scriptsize{\textcolor{gray!90}{low}}};
\node (medium) at (0,-0.3) {\scriptsize{\textcolor{gray!90}{medium}}};
\node (high) at (3.6,-0.3) {\scriptsize{\textcolor{gray!90}{high}}};
\end{tikzpicture}
\caption{Overlay of all edges and items used in an exemplary final population $P_2$ on instance~$1$ with $\alpha = 10\%$ (left) and $\alpha = 50\%$ (right). Edges and items are coloured by their frequency.}
\label{fig:frq+map2}
\end{figure*}
\section{Experimental Investigation}
\label{Sec:EXP}

We empirically study the Co-EA in this section. We run the Co-EA on eighteen TTP instances from~\cite{PolyakovskiyB0MN14}, the same instances are used in~\cite{NikfarjamMap}. We first illustrate the distribution of solutions in $P_1$, and the structural diversity of solutions in $P_2$. Then, we compare the self-adaptation methods with the fixed parameter setting. Afterwards, we conduct a comprehensive comparison between $P_1$ and $P_2$ and the populations obtained by~\cite{NikfarjamMap} and~\cite{NikTTPEDO}. Here, the termination criterion and $\alpha$ are set on $1000000m$ fitness evaluations and $10\%$, respectively. 

MAP-Elite selection can be beneficial to illustrate the distribution of high-quality solutions in the behaviour space. On the other hand, EDO selection aims to understand which elements in high-quality solutions is easy/difficult to be replaced. Figure \ref{fig:frq+map} depicts exemplary populations $P_1$ and $P_2$ after $4000m$ and $1000000m$ fitness evaluations of Co-EA on instance~$1$, where $\alpha = 2\%$. The first row illustrates the distribution $P_1$'s high-performing solutions over the behavioural space of $f(x)$ and $g(y)$. The second and the third rows represent the overlay of edges and items in $P_2$, respectively. The figure shows the solutions with highest quality are located on top-right of the map on this test instance where the gaps of $f(x)$ and $g(y)$ to $f^*$ and $g^*$ are in $[0.015 0.035]$ and $[0.15 0.18]$, respectively. In the second row of the figure, we can observe that Co-EA successfully incorporates new edges into $P_2$ and reduces the edges' frequency within the population. However, it is unsuccessful in incorporating new items in $P_2$. The reason can be that there is a strong correlation between items in this particular test instance, and the difference in the weight and profit of items is significant.  It means that there is not many other good items to be replaced with the current selection. Thus, we cannot change the items easily when the quality criterion is fairly tight ($\alpha = 2\%$). As shown on the third row of the figure, the algorithm can change $i_8$ with $i_{43}$ in some packing lists.

\begin{table*}
\centering
\caption{Comparison of $Gamma_1$ (1) and  $Gamma_2$ (2), and the fixed method~(3). The instances are numbered as Table~1 in~\cite{NikfarjamMap}. In columns Stat the notation $X^+$ means the median of the measure is better than the one for variant $X$, $X^-$ means it is worse, and $X^*$ indicates no significant difference. Stat shows the results of Kruskal-Wallis statistical test at a significance level of $5\%$ and Bonferroni correction. Also, $p^* = \max_{p\in P_1} \{z(p)\}$.}
\centering
\renewcommand{\tabcolsep}{3pt}
\renewcommand{\arraystretch}{1.2}
\begin{tabular}{r|cccccc|cccccc}
\toprule
             &&&$H(P_2)$&&&&&&$z(p^*)$&&&\\
\cmidrule(l{2pt}r{2pt}){2-7}
\cmidrule(l{2pt}r{2pt}){8-13}
             Inst. & $Gamma_1$ &(1) & $Gamma_2$& (2) &$fixed$ & (3) & $Gamma_1$ &(1) & $Gamma_2$& (2) &$fixed$ & (3)\\
\cmidrule(l{2pt}r{2pt}){2-3}
\cmidrule(l{2pt}r{2pt}){4-5}
\cmidrule(l{2pt}r{2pt}){6-7}
\cmidrule(l{2pt}r{2pt}){8-9}
\cmidrule(l{2pt}r{2pt}){10-11}
\cmidrule(l{2pt}r{2pt}){12-13}
            & mean  &Stat & mean &Stat & mean &Stat& mean  &Stat & mean &Stat & mean &Stat\\
\midrule
1&8.7&$2^-3^+$&\hl{8.8}&$1^+3^+$&8.2&$1^-2^-$&4452.4&$2^-3^*$&\hl{4465}&$1^+3^*$&4461.1&$1^*2^*$\\
2&9.3&$2^*3^+$&9.3&$1^*3^+$&9.1&$1^-2^-$&\hl{8270.4}&$2^*3^*$&8232.2&$1^*3^*$&8225.2&$1^*2^*$\\
3&\hl{9.9}&$2^+3^+$&9.8&$1^-3^+$&9.6&$1^-2^-$&13545.4&$2^*3^*$&13607.5&$1^*3^*$&\hl{13609}&$1^*2^*$\\
4&7.7&$2^-3^+$&7.7&$1^+3^+$&7.4&$1^-2^-$&1607.1&$2^*3^*$&1607.5&$1^*3^*$&1607.5&$1^*2^*$\\
5&9&$2^*3^+$&9&$1^*3^+$&8.8&$1^-2^-$&\hl{4814.7}&$2^*3^*$&4805.3&$1^*3^*$&4811&$1^*2^*$\\
6&9.4&$2^*3^+$&9.4&$1^*3^+$&9.2&$1^-2^-$&6834.5&$2^*3^*$&6850&$1^*3^*$&6850&$1^*2^*$\\
7&8&$2^*3^+$&\hl{8.1}&$1^*3^+$&7.6&$1^-2^-$&3200.8&$2^*3^*$&\hl{3218.4}&$1^*3^*$&3165&$1^*2^*$\\
8&9&$2^-3^+$&9&$1^+3^+$&8.8&$1^-2^-$&7854.2&$2^*3^*$&7854.2&$1^*3^*$&7850.9&$1^*2^*$\\
9&9.5&$2^*3^+$&9.5&$1^*3^+$&9.3&$1^-2^-$&13644.8&$2^+3^+$&13644.8&$1^-3^+$&13644.8&$1^-2^-$\\
10&10.5&$2^-3^+$&10.5&$1^+3^+$&10.1&$1^-2^-$&11113.6&$2^*3^*$&11145.7&$1^*3^-$&\hl{11148}&$1^*2^+$\\
11&11.2&$2^*3^+$&11.2&$1^*3^+$&11&$1^-2^-$&25384.6&$2^+3^*$&\hl{25416.6}&$1^-3^-$&25401.3&$1^*2^+$\\
12&9.3&$2^*3^+$&\hl{9.4}&$1^*3^+$&9.2&$1^-2^-$&3538.2&$2^*3^*$&\hl{3564.4}&$1^*3^*$&3489.4&$1^*2^*$\\
13&10.7&$2^*3^+$&10.7&$1^*3^+$&10.5&$1^-2^-$&\hl{13369.3}&$2^+3^*$&13310.4&$1^-3^*$&13338.4&$1^*2^*$\\
14&7.7&$2^-3^*$&9.7&$1^+3^+$&8.5&$1^*2^-$&5261.9&$2^*3^*$&\hl{5410.3}&$1^*3^*$&5367.1&$1^*2^*$\\
15&10.9&$2^*3^+$&10.9&$1^*3^+$&10.7&$1^-2^-$&20506.8&$2^*3^*$&20506.8&$1^*3^*$&20385.3&$1^*2^*$\\
16&11.6&$2^-3^*$&\hl{11.7}&$1^+3^+$&11.4&$1^*2^-$&18622.2&$2^*3^*$&18609.6&$1^*3^*$&\hl{18641.4}&$1^*2^*$\\
17&11.2&$2^*3^+$&11.2&$1^*3^+$&11.1&$1^-2^-$&9403.8&$2^*3^*$&\hl{9448.3}&$1^*3^*$&9428.1&$1^*2^*$\\
18&11.4&$2^*3^+$&11.4&$1^*3^+$&11.1&$1^-2^-$&19855.3&$2^-3^*$&\hl{19943.8}&$1^+3^*$&19879.3&$1^*2^*$\\
\bottomrule
\end{tabular}
\label{tbl:Res_adpt}
\end{table*}
Figure~\ref{fig:frq+map2} reveals that, as $\alpha$ increases, so does the room to involve more items and edges in $P_2$. In other words, there can be found more edges and items to be included in $P_2$. Figure~\ref{fig:frq+map2} shows the overlays on the same instances, where $\alpha$ is set to $10\%$~(left) and $50\%$~(right). Not only more edges and items are included in $P_2$ with the increase of $\alpha$, but also Co-EA reduces the frequency of the edges and items in $P_2$ to such a degree that we can barely see any high-frequent edges or items in the figures associated with $\alpha = 50\%$. Moreover, the algorithm can successfully include almost all items in $P_2$ except item $i_{39}$. Checking the item's weight, we notice that it is impossible to incorporate the item into any solution. This is because, $w_{i_{39}} = 4\,400$, while the capacity of the knapsack is set to $4\,029$. In other words, $w_{i_{39}} > W$.

\subsection{Analysis of Self-Adaptation}
In this sub-section, we compare the two proposed termination criteria and self-adaptation methods $Gamma_1$ and $Gamma_2$ with the fixed method employed in~\cite{NikfarjamMap}. We incorporate these methods into the Co-EA and run it for ten independent runs. Table~\ref{tbl:Res_adpt} summarises the mean of $P_2$'s entropy obtained from the competitors. The table indicates that both $Gamma_1$ and $Gamma_2$ outperform the fixed method on all test instances. Kruskal-Wallis statistical tests at significance level $5\%$ and Bonferroni correction also confirm a meaningful difference in median of results for all instances except instance 15 where there is no significant difference in the mean of $Gamma_1$ and the fixed method. In comparison between $Gamma_1$ and $Gamma_2$, the latter outperforms the first in $4$ test instances, while it is surpassed in only one case. In conclusion, Table~\ref{tbl:Res_adpt} indicates that $Gamma_2$ works the best with respect to the entropy of $P_2$.

\begin{table*}
\centering
\caption{Comparison of the Co-EA and QD from \cite{NikfarjamMap} in terms of $z(p^*)$, and EDO algorithm from \cite{NikTTPEDO} in $H(P_2)$. Stat shows the results of Mann-Whitney U-test at significance level $5\%$. The notations are in line with Table \ref{tbl:Res_adpt}.}
\renewcommand{\tabcolsep}{9pt}
\renewcommand{\arraystretch}{1}
\begin{tabular}{r|cccc|cccc}
\toprule
             Inst. & Co-EA &(1) & QD &(2) & Co-EA &(1) & EDO& (2)\\
\cmidrule(l{2pt}r{2pt}){2-3}
\cmidrule(l{2pt}r{2pt}){4-5}
\cmidrule(l{2pt}r{2pt}){6-7}
\cmidrule(l{2pt}r{2pt}){8-9}
            & $Q$  & Stat &$Q$ & Stat  & $H$ &Stat & $H$  &Stat\\
\midrule
1&\hl{4465}&$2^*$&4463.5&$1^*$&\hl{8.8}&$2^*$&8.6&$1^*$\\
2&\hl{8232.2}&$2^*$&8225.7&$1^*$&9.3&$2^-$&\hl{9.4}&$1^+$\\
3&\hl{13607.5}&$2^*$&13544.9&$1^*$&9.8&$2^-$&9.8&$1^+$\\
4&1607.5&$2^+$&1607.5&$1^-$&7.7&$2^+$&7.7&$1^-$\\
5&4805.3&$2^*$&\hl{4813.2}&$1^*$&9&$2^*$&9&$1^*$\\
6&\hl{6850}&$2^*$&6806.8&$1^*$&\hl{9.4}&$2^+$&9.3&$1^-$\\
7&\hl{3218.4}&$2^*$&3191.9&$1^*$&\hl{8.1}&$2^+$&8&$1^-$\\
8&\hl{7854.2}&$2^*$&7850.9&$1^*$&9&$2^*$&9&$1^*$\\
9&13644.8&$2^+$&13644.8&$1^-$&9.5&$2^*$&9.5&$1^*$\\
10&11145.7&$2^-$&\hl{11149.2}&$1^+$&\hl{10.5}&$2^+$&10.2&$1^-$\\
11&25416.6&$2^-$&\hl{25555.2}&$1^+$&\hl{11.2}&$2^+$&11&$1^-$\\
12&\hl{3564.4}&$2^*$&3514&$1^*$&\hl{9.4}&$2^+$&8.8&$1^-$\\
13&13310.4&$2^*$&\hl{13338.6}&$1^*$&\hl{10.7}&$2^+$&10.2&$1^-$\\
14&\hl{5410.3}&$2^*$&5364.6&$1^*$&\hl{9.7}&$2^+$&9.5&$1^-$\\
15&\hl{20506.8}&$2^*$&20499.2&$1^*$&\hl{10.9}&$2^+$&10.7&$1^-$\\
16&18609.6&$2^-$&\hl{18666.4}&$1^+$&\hl{11.7}&$2^+$&11.1&$1^-$\\
17&\hl{9448.3}&$2^*$&9407.7&$1^*$&\hl{11.2}&$2^+$&10.4&$1^-$\\
18&\hl{19943.8}&$2^+$&19861.8&$1^-$&\hl{11.4}&$2^+$&11.1&$1^-$\\
\bottomrule
\end{tabular}
\label{tbl:Res_OPR}
\end{table*}
\begin{figure*}
\centering
\includegraphics[width=0.9\columnwidth]{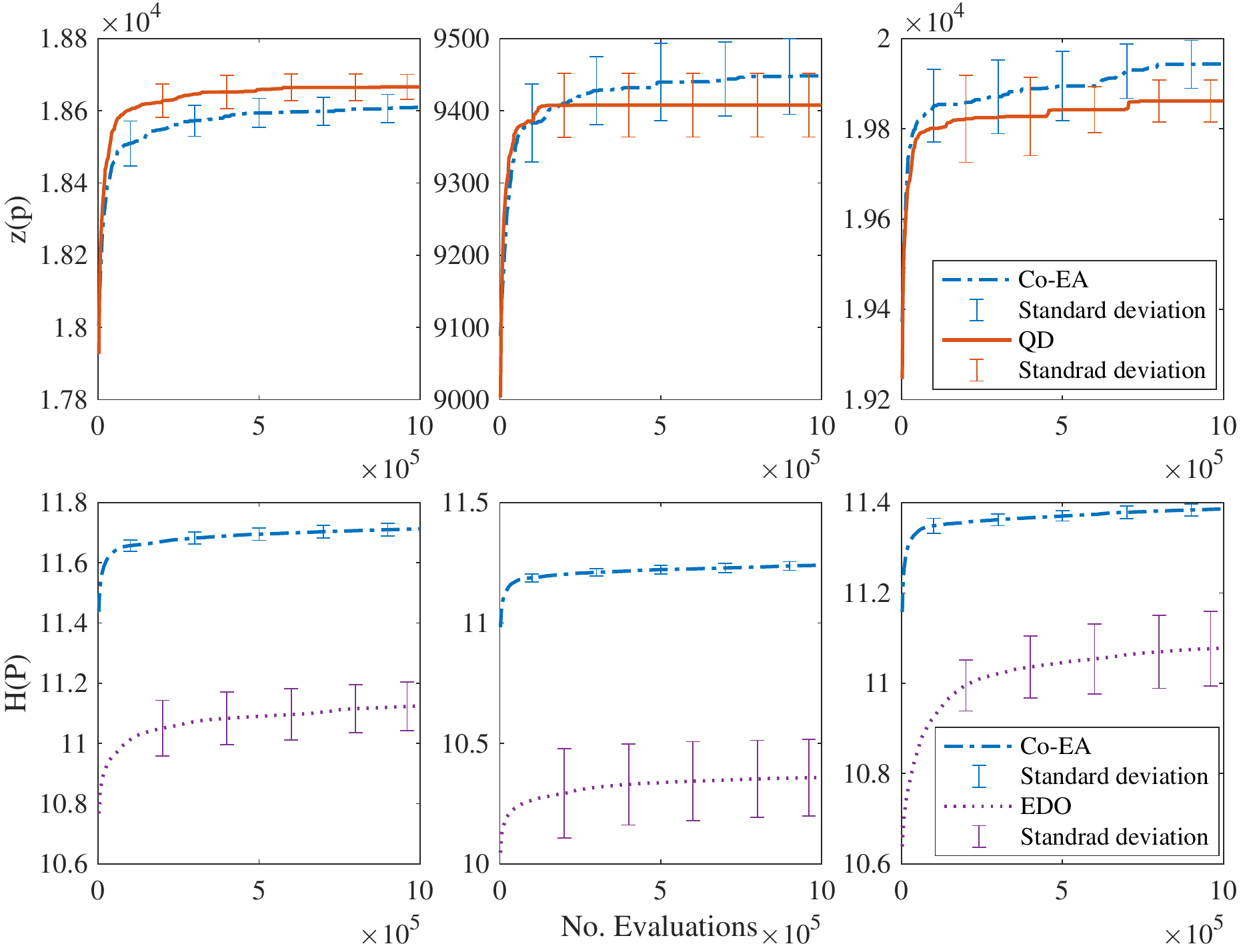}

\caption{Representative trajectories of Co-EA and standard EDO~EA on instances 16, 17, 18. The top row shows $H(p_2)$ while the second row shows the best solution in $P_1$.}
\label{fig:Trd}
\end{figure*}

Moreover, Table~\ref{tbl:Res_adpt} also shows the mean TTP score of the best solution in $P_1$ obtained from the three competitors. Although Table~\ref{tbl:Res_adpt} indicates that the statistical test cannot confirm a significant difference in the mean of the best TTP solutions, $Gamma_2$'s results are slightly better in~$7$ cases, while $Gamma_1$ and $fixed$ have better results in~$3$ cases. Overall, all three competitors perform almost equally in terms of the best TTP score. Since $Gamma_2$ outperforms other methods in entropy, we employ it for the Co-EA in the rest of the study.

\subsection{Analysis of Co-EA}
This section compares $P_1$ and $P_2$ with the QD-based EA in~\cite{NikfarjamMap} and the standard EDO algorithm, respectively. Table~\ref{tbl:Res_OPR} summarises this series of experiments. The results indicate that the Co-EA outperforms the standard EDO in 14 instances, while the EDO algorithm has a higher entropy average in only two cases. In the two other test instances, both algorithms performed equally. Moreover, the Co-EA yields competitive results in terms of the quality of the best solution compared to the QD-based EA; in fact, the Co-EA results in a higher mean of TTP scores on 12 test instances. For example, the best solutions found by Co-EA score $19943.8$ on average, whereby the figure stands at $19861.8$ for the QD-based algorithms.     

Figure~\ref{fig:Trd} depicts the trajectories of Co-EA and the standard EDO algorithm in entropy of the population (the first row), and that of Co-EA and QD-based EA in quality of the best solution (the second row). Note that in the first row the $x$-axis shows fitness evaluations from $4000m$ to $1000000m$. This is because $P_2$ is empty in the early stages of running Co-EA and we cannot calculate the entropy of $P_2$ until $|P_2| = \mu$ for the sake of fair comparison. The figure shows that Co-EA converges faster and to a higher entropy than the standard EDO algorithm. Moreover, it also depicts results obtained by Co-EA has much less standard deviation. Regarding the quality of the best solution, both Co-EA and QD-based EA follow a similar trend.

\section{Conclusion}
We introduced a co-evolutionary algorithm to simultaneously evolve two populations for the traveling thief problem. The first population explore niches in a behavioural space and the other maximises structural diversity. The results showed superiority of the algorithm to the standard framework in the literature in maximising diversity. The co-evolutionary algorithm also yields competitive results in terms of quality. 

It is intriguing to adopt more complicated MAP-Elites-based survival selection for exploring the behavioural space. Moreover, this study can be a transition from benchmark problems to real-world optimisation problems where imperfect modelling is common and diversity in solutions can be beneficial.

\section*{Acknowledgements}
This work was supported by the Australian Research Council through grants DP190103894 and FT200100536.
\bibliographystyle{abbrvnat}
\bibliography{Main}

\end{document}